\documentclass[11pt]{article}
\usepackage{coling2020}
\usepackage{times}
\usepackage{url}
\usepackage{latexsym}

\usepackage{graphicx}  
\usepackage{url}
\usepackage{graphicx}
\usepackage{float} 
\usepackage{subfigure}
\usepackage{amsmath}
\usepackage{amsfonts}
\usepackage{multirow}
\usepackage{array}
\usepackage{makecell}
\usepackage[utf8]{inputenc}
\usepackage{enumerate}

\usepackage{xcolor}
\usepackage{soul}
\newcolumntype{L}[1]{>{\raggedright\let\newline\\\arraybackslash\hspace{0pt}}m{#1}}
\setul{2pt}{2pt}
\usepackage{multirow, booktabs}

\usepackage{pifont}
\newcommand{\xmark}{\ding{55}}
\usepackage{microtype}

\title{A Dependency Syntactic Knowledge Augmented Interactive Architecture \\ for End-to-End Aspect-based Sentiment Analysis}

\author{
  Yunlong Liang\textsuperscript{1}\thanks{ \ \ Work was done when Yunlong Liang was an intern at Pattern Recognition Center, WeChat AI, Tencent Inc, China.}  , 
  Fandong Meng\textsuperscript{2}, 
  Jinchao Zhang\textsuperscript{2}, 
  Jinan Xu\textsuperscript{1}\thanks{ \ \ Jinan Xu is the corresponding author.}\\
  \textbf{Yufeng Chen}\textsuperscript{1} and
  \textbf{Jie Zhou}\textsuperscript{2}\\
  \textsuperscript{1}Beijing Jiaotong University, China \\
  \textsuperscript{2}Pattern Recognition Center, WeChat AI, Tencent Inc, China \\
  \texttt{\{yunlonliang,jaxu,chenyf\}@bjtu.edu.cn} \\
  \texttt{\{fandongmeng,dayerzhang,withtomzhou\}@tencent.com} \\
}
\date{}

\begin{document}
\maketitle
\begin{abstract}
  The aspect-based sentiment analysis (ABSA) task remains to be a long-standing challenge, which aims to extract the aspect term and then identify its sentiment orientation. 
  In previous approaches, the explicit syntactic structure of a sentence, which reflects the syntax properties of natural language and hence is intuitively crucial for aspect term extraction and sentiment recognition, is typically neglected or insufficiently modeled. 
  In this paper, we thus propose a novel dependency syntactic knowledge augmented interactive architecture with multi-task learning for end-to-end ABSA. 
  This model is capable of fully exploiting the syntactic knowledge (dependency relations and types) by leveraging a well-designed \textbf{D}ependency \textbf{R}elation \textbf{E}mbedded \textbf{G}raph \textbf{C}onvolutional \textbf{N}etwork (\textsc{DreGcn}).
  Additionally, we design a simple yet effective message-passing mechanism to ensure that our model learns from multiple related tasks in a multi-task learning framework. 
  Extensive experimental results on three benchmark datasets demonstrate the effectiveness of our approach, which significantly outperforms existing state-of-the-art methods. 
  Besides, we achieve further improvements by using BERT as an additional feature extractor.
\end{abstract}

\section{Introduction} \label{sec:introduction}
\begin{figure*}[!hbt]
    \centering
    \includegraphics[width=0.5\textwidth]{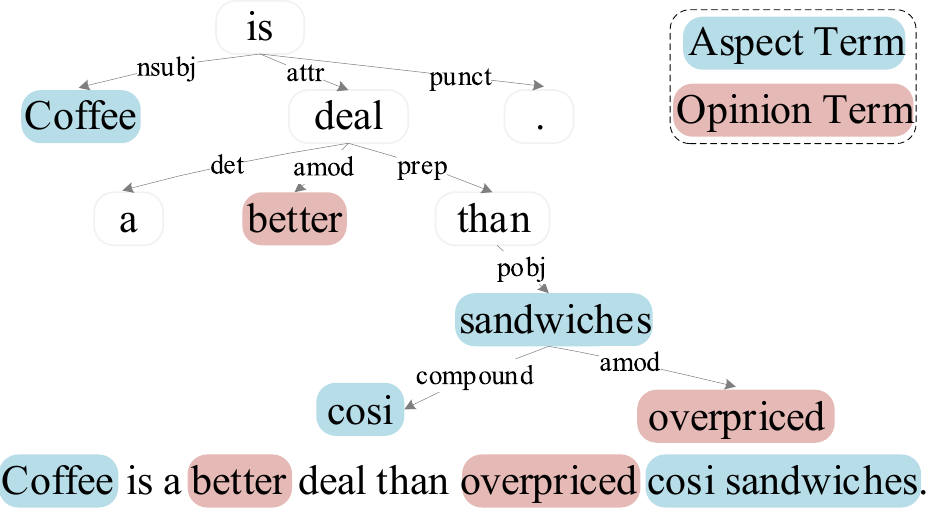}
    \caption{An example of dependency tree (generated by spaCy~\cite{Dependency_tree}). For instance, the dependency relation $sandwiches\xrightarrow{compound}cosi$ means ``\emph{sandwiches}'' is a nominal modifier of ``\emph{cosi}''. The tree can be easily converted into a dependency graph representation where words are regarded as nodes, and dependency relation types become edges. }
    \label{fig:case}
\end{figure*}

The aspect-based sentiment analysis (ABSA) is a long-challenging task, which consists of two subtasks: \textbf{a}spect term \textbf{e}xtraction (AE) and \textbf{a}spect-level \textbf{s}entiment classification (AS). The AE task aims to extract aspect terms from the given text. The goal of the AS task is to detect the sentiment orientation over the extracted aspect terms. For example, in Figure~\ref{fig:case}, 
there are two aspect terms mentioned in the sentence, namely, ``\emph{\textbf{coffee}}'' and ``\emph{\textbf{cosi sandwiches}}'', towards which the sentiment polarity is {\em positive} and {\em negative}, respectively. 


For the overall ABSA task, previous work has shown that the joint approaches~\cite{he_acl2019,Luo2019doer} can achieve better results than pipeline or integrated methods~\cite{Wang18,li2019unified}, since the joint approaches can sufficiently model the correlation between the two subtasks, i.e., AE and AS. However, these models are typically insufficient for modeling the syntax information, which reveals internal logical relations between words and thus is intuitively pivotal to the ABSA task. For instance, in Figure~\ref{fig:case}, 1) given ``\emph{\textbf{sandwiches}}'' as a part of an aspect term, ``\emph{\textbf{cosi}}'' can also be extracted as a part of the aspect term through the dependency relation type {\em compound} with ``\emph{\textbf{sandwiches}}'', and thus constitutes a complete aspect term with ``\emph{\textbf{sandwiches}}'', namely, ``\emph{\textbf{cosi sandwiches}}''; 2) after the aspect term being extracted, the sentiment polarity of ``\emph{\textbf{cosi sandwiches}}'' can be easily classified as {\em negative} due to the opinion word ``\emph{overpriced}'', which is pointed out by the dependency relation: $sandwiches\xrightarrow{amod}overpriced$ ($amod$ means adjectival modifier); 3) for the sentiment orientation of another aspect term, the opinion word ``\emph{better}'' indicates that the sentiment polarity of ``\emph{\textbf{coffee}}'' is {\em positive} through the multi-level dependency relation: $coffee\xrightarrow{nsubj}is\xrightarrow{attr}deal\xrightarrow{amod}better$. Clearly, for the sentiment orientation of multiple aspects, the model will be not confused if differentiated dependency relation paths are considered appropriately. Therefore, a syntax-independent encoder may not encode such critical relational information into the final representation, which may lead to incorrect predictions.

Recent studies with separate subtask settings have indeed shown that the syntax information can benefit the AE task and the 
AS task. 
For instance, for the AE task, Dai and Song~\shortcite{dai-song-2019-neural} manually design some aspect term extraction patterns based on a few dependency relations, and then construct a large amount of auxiliary data to improve the performance. To enhance the tree-structured representation to improve the AE performance, Luo et al.~\shortcite{Luo2019_tree} encode the dependency relation as features by using a bidirectional gate control mechanism in dependency trees, which originates from bidirectional LSTM~\cite{Hochreiter:1997:LSM:1246443.1246450}.
For the AS task, the variants of recursive neural network~\cite{socher2011parsing} or graph convolutional network (GCN,~\cite{kipf2017semi}) are exploited to capture the syntactic information from dependency (constituent) tree of the sentence to make the representation of the target aspect richer for more accurate sentiment predictions~\cite{dong-etal-2014-adaptive,nguyen-shirai-2015-phrasernn,huang-carley-2019-syntax,zhang-etal-2019-aspect}. 
Wang et al.~\shortcite{8561296} utilize a syntax-directed local attention to lay more emphasis on the words syntactically close to the target aspect instead of the position-based ones for more performance gains of the AS task. 
However, these studies do not simultaneously enhance the two subtasks with syntactic knowledge in a joint framework, which is beneficial to each subtask and thus can improve the overall performance of the ABSA. Additionally, dependency relation types are not sufficiently exploited to improve the overall performance. 

Therefore, we propose a dependency syntactic knowledge augmented interactive architecture with multi-task learning, which is able to fully exploit the syntactic knowledge and simultaneously model multiple related tasks. In particular, we design a \textbf{D}ependency \textbf{R}elation \textbf{E}mbedded \textbf{G}raph \textbf{C}onvolutional \textbf{N}etwork (\textsc{DreGcn}) to fully model the dependency relation as well as the dependency relation type between words in one sentence. Furthermore, we propose a simple yet more effective message-passing mechanism to ensure that our model learns from multiple different but related tasks.

We evaluate our approach on three benchmark datasets. Experimental results demonstrate the effectiveness of our model, which significantly outperforms existing systems and achieves new state-of-the-art performance. Besides, we provide further improvements by using BERT~\cite{bert} as an additional feature extractor. 

Our contributions can be summarized as follows:
\begin{itemize}
\item We propose a novel \textbf{D}ependency \textbf{R}elation \textbf{E}mbedded \textbf{G}raph \textbf{C}onvolutional \textbf{N}etwork (\textsc{DreGcn}) for the overall ABSA task in a joint framework, which is capable of fully exploiting the more fine-grained linguistic knowledge (e.g., the dependency relation and type) at the relational level than vanilla GCN. 
\item We propose a more effective message-passing mechanism to ensure the model learns from multiple related tasks.
\item Our approach substantially outperforms previous systems and achieves consistently state-of-the-art results on three benchmark datasets.

\end{itemize}


\begin{figure}[!hbt]
    \centering
    \includegraphics[width=0.48\textwidth]{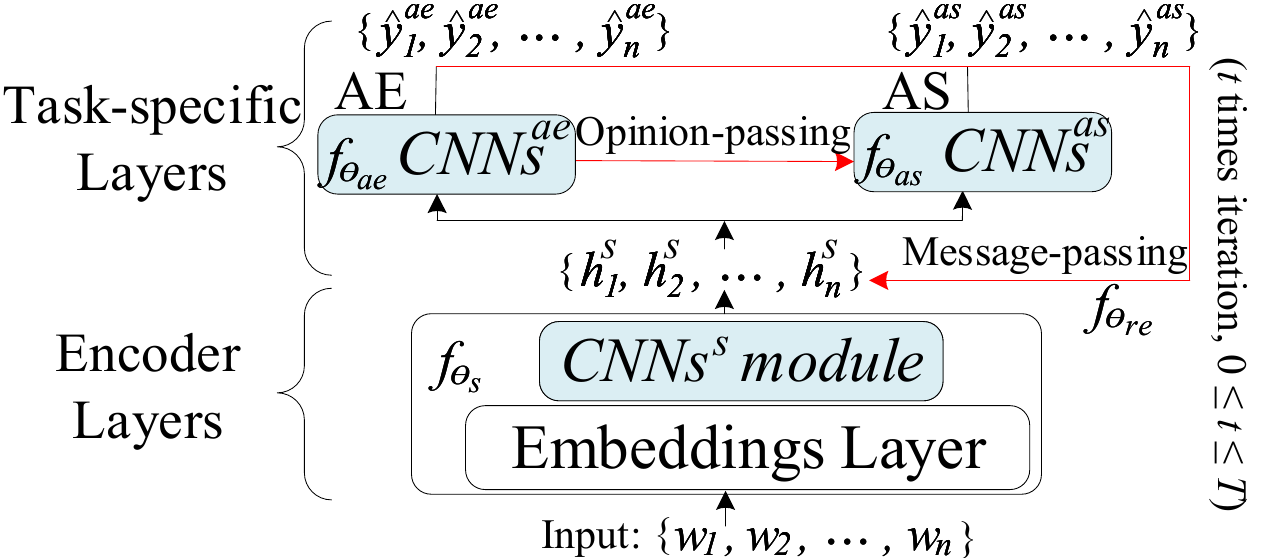}
    \caption{Overview of the interactive architecture. ``$t$'' denotes the iteration number and ``$T$'' denotes the maximum number of iterations in the message-passing mechanism. Document-level parts are removed compared with the original work~\cite{he_acl2019}. 
    }
    \label{fig:cnn_origin}
\end{figure}

\section{Background}

\subsection{Task Definition}

We formulate the complete aspect-based sentiment analysis (ABSA) task as two sequence labeling subtasks, namely, \textbf{a}spect term \textbf{e}xtraction (AE)\footnote{Aspect and opinion term co-extraction are simultaneously performed. In this paper, AE denotes these two tasks for simplicity. } and \textbf{a}spect-level \textbf{s}entiment classification (AS). For the AE task, following~\cite{he_acl2019}, we employ the BIO tagging scheme: $\mathcal{Y}^{ae} = \{\texttt{BA},\texttt{IA},\texttt{BP}, \texttt{IP}, \texttt{O}\}$ to label all the aspect and opinion terms mentioned in the sentence\footnote{A word can not belong to both aspect term and opinion term at the same time.}. $\texttt{BA}$ and $\texttt{IA}$ denotes the beginning and inside of an aspect term, respectively. $\texttt{BP}$ and $\texttt{IP}$ denotes the beginning and inside of an opinion term, respectively, and $\texttt{O}$ denotes other words. For the AS task, we employ the label set: $\mathcal{Y}^{as} = \{\texttt{pos},\texttt{neg},\texttt{neu}\}$ to mark the token-level sentiment polarity. $\texttt{pos},\texttt{neg}$ and $\texttt{neu}$ indicates the {\em positive}, {\em negative} and {\em neutral} sentiment polarity, respectively. 
Given an input sentence $\mathrm{ X}=\{w_1,w_2,\dots,w_n\}$ with length $n$, our goal is to predict two tag sequences $\mathrm{ Y}^{ae}=\{{y}_1^{ae},{y}_2^{ae},\dots,{y}_n^{ae}\}$ and $\mathrm{ Y}^{as}=\{{y}_1^{as},{y}_2^{as},\dots,{y}_n^{as}\}$, where $y^{{ae}}_i \in \mathcal{Y}^{{ae}}$, $y^{{as}}_i \in \mathcal{Y}^{{as}}$, respectively, $1 \leq i \leq n$. 

\subsection{An Interactive Architecture with Multi-task Learning}
\label{sec:interactive_architecture}
Figure~\ref{fig:cnn_origin} is the interactive architecture with multi-task learning, proposed by~\cite{he_acl2019}, 
which is the current state-of-the-art model for the end-to-end ABSA task\footnote{Peng et al.~\shortcite{peng2019knowing} also achieve good performances on the end-to-end ABSA task but they focus on the limited scenario where the opinion term and corresponding aspect term need to be paired in one sentence.}, 
in which the Encoder Layers encode the sentence representation for multiple related tasks. The Task-specific Layers, which consist of two key components: message-passing and opinion-passing mechanism, serves as predictions for different tasks. 

For an input sequence, the feature extractor $f_{\theta _s}$ maps the input to a shared latent sequence $\{\mathbf{h}_1^s,\mathbf{h}_2^s,\dots,\mathbf{h}_n^s\}$\footnote{The iteration superscript $t$ in the description is omitted for simplicity, i.e., $\{\mathbf{h}_1^s,\mathbf{h}_2^s,\dots,\mathbf{h}_n^s\}$ = $\{\mathbf{h}_1^{s(t)},\mathbf{h}_2^{s(t)},\dots,\mathbf{h}_n^{s(t)},  t=0 \}$.}. Then task-specific component AE assigns to each token with a probability distribution:
$\mathbf{\hat{y}}_1^{ae},\mathbf{\hat{y}}_2^{ae},\dots,\mathbf{\hat{y}}_n^{ae} = f_{\theta _{ae}}(\mathbf{h}_1^{s},\mathbf{h}_2^{s},\dots,\mathbf{h}_n^{s})$, where the top value of the probability distribution of each token indicates whether it is a part of any aspect terms or opinion terms. 
The output of the AS component is formulated as: 
$\mathbf{\hat{y}}_1^{as},\mathbf{\hat{y}}_2^{as},\dots,\mathbf{\hat{y}}_n^{as} = f_{\theta _{as}}(\mathbf{h}_1^{s},\mathbf{h}_2^{s},\dots,\mathbf{h}_n^{s})
$. Then, message-passing mechanism will update the sequence of shared latent vectors by combining the probability distribution of the AE and AS task:
\begin{equation}
\setlength{\abovedisplayskip}{5pt}
\setlength{\belowdisplayskip}{5pt}
  \label{eq:message-passing-predictions}
  \begin{split}
   \mathbf{h}_i^{s(t)} &= f_{\theta _{re}}(\mathbf{h}_i^{s(t-1)};\mathbf{\hat{y}}_i^{ae(t-1)};\mathbf{\hat{y}}_i^{as(t-1)})
  \end{split}
\end{equation}
where $\mathbf{h}_i^{s(t)}$ denotes the shared latent vector corresponding to $w_i$ after $t$ rounds of message-passing; $f_{\theta _{re}}$ is a re-encoding function (i.e. fully-connected layer) and [;] means concatenation. 

Meanwhile, the opinion information from the AE task is sent to the AS task as shown in Figure~\ref{fig:cnn_origin}, which is useful to the AS task. Specifically, a self-attention matrix $\mathbf{M} \in \mathbb{M}^{n \times n}$ is employed:
\begin{equation}
\setlength{\abovedisplayskip}{5pt}
\setlength{\belowdisplayskip}{5pt}
\label{revelance_fun}
    \mathbf{S}_{ij}^{(i \neq j)} = (\mathbf{h}_i^{as} \mathbf{W}_s (\mathbf{h}_j^{as})^T) \cdot \frac{1}{|i-j|} \cdot P_j^{op} ;\qquad
    \mathbf{M}_{ij}^{(i \neq j)} = \frac{\text{exp}(\mathbf{S}_{ij})}{\sum_{k=1}^n \text{exp}(\mathbf{S}_{ik})} 
\end{equation}
where $i \neq j$ means we only consider context words for inferring the sentiment
of the target token; $\mathbf{W}_s$ is the transformation matrix; $\frac{1}{|i-j|}$ is a distance-related factor and $P_j^{op}$ is computed by summing the predicted probabilities of $\mathbf{y}_j^{ae}$ which is the predicted probability on opinion-related labels (i.e. BP and IP). The Eq.(\ref{revelance_fun}) aims to measure the semantic relevance between $\mathbf{h}_i^{as}$ and $\mathbf{h}_j^{as}$. Finally, $\mathbf{h}_i^{as}$ and $\mathbf{h}_i^{\prime as}$ are concatenated as the output representation of the AS part where $\mathbf{h}_i^{\prime as} = \sum_{j=1}^n \mathbf{M}_{ij} \mathbf{h}_j^{as}$. 

Although the interactive architecture mentioned above has achieved state-of-the-art performance, there still exist two drawbacks: 1) the architecture neglects the syntax modeling; and 2) the probability distribution is insufficient to pass the rich task-specific information. We thus propose a syntax augmented interactive architecture, which can fully exploit the syntax information by utilizing a dependency relation embedded graph convolutional network (\textsc{DreGcn}). And we also design a more effective message-passing mechanism. The whole model with those two key components will be elaborated in the next section. 

\begin{figure*}
    \centering
    \includegraphics[width=0.96\textwidth]{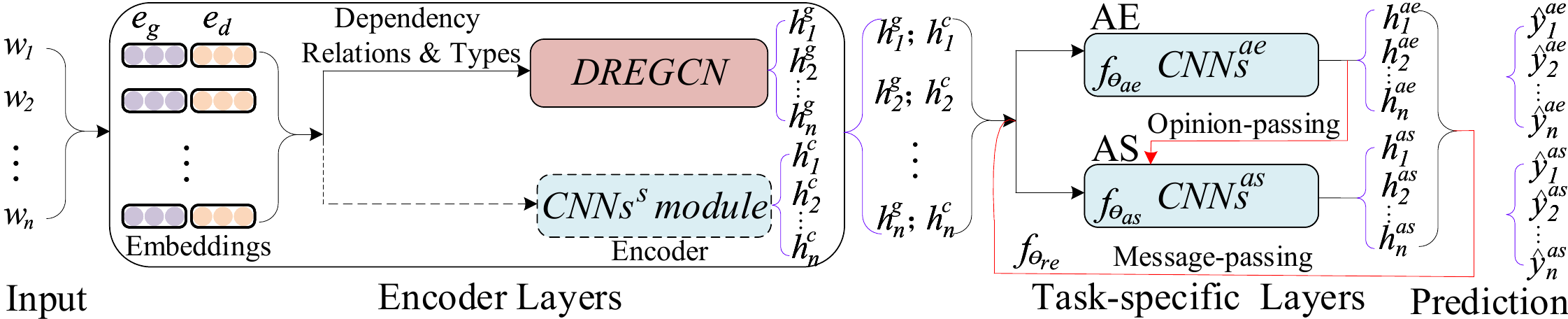}
    \caption{Architecture of the proposed approach. $e_g$ and $e_d$ mean general-purpose and domain-specific embeddings, respectively. [;] denotes concatenation. AE: aspect term and opinion term co-extraction; AS: aspect-level sentiment classification.}
    \label{fig:architecture}
\end{figure*}
\section{Approach}
\subsection{Overview}
In Figure~\ref{fig:architecture}, from left to right, our approach has two key components, described in detail with the callouts, to investigate two important intuitions in the ABSA task. Firstly, we carefully design a dependency relation embedded graph convolutional network (\textsc{DreGcn}) in the Encoder Layers, which aim to fully exploit the syntactic knowledge. Secondly, we propose a more effective message-passing mechanism in the Task-specific Layers to make the model learn from multiple related tasks. 


\subsection{Encoder Layers}
To exploit the syntactic knowledge, we design a dependency relation embedded graph convolutional network (GCN,~\cite{kipf2017semi}) in the Encoder Layers. Additionally, we retain the convolutional neural network (CNN) as an alternative part, because the n-gram features at different granularities are important to the ABSA task. 

GCN aggregates the feature vectors of neighboring nodes and propagates the information of a node to its first-order neighbors. For a dependency tree with $n$ nodes, an $n \times n$ adjacency matrix $\mathbf{A}$ can be generated. As done in~\cite{schlichtkrull2018modeling}, we add a self-loop for each node and include the reversed direction of a dependency arc if there is a dependency relation between node $i$ and node $j$, i.e., $\mathbf{A}_{ij}$ = $\mathbf{A}_{ji}$ = 1, otherwise $\mathbf{A}_{ij}$ = $\mathbf{A}_{ji}$ = 0. 
Then a GCN layer can obtain new node features by convolving the neighboring nodes' features by the following function:
\begin{equation}
\setlength{\abovedisplayskip}{5pt}
\setlength{\belowdisplayskip}{5pt}
  \label{gcn}
  \begin{split}
  \mathbf{h}^{l+1}_{i} &= ReLU ( \sum_{j=1}^{n}(\mathbf{A}_{ij}\mathbf{W}^{l+1}_g\mathbf{h}^{l}_{j}+\mathbf{b}^{l+1}_g))
  \end{split}
\end{equation}
where $i$ is the current node and $j$ denotes the neighborhood of node $i$; $\mathbf{h}^{l}_{j}$ represents the feature of node $j$ at layer $l$; $\mathbf{W}_g$ and $\mathbf{b}_g$ are trainable weights, mapping the feature of a node to its adjacent nodes in the graph; $\mathbf{h}, \mathbf{b}_g \in \mathbb{R}^{d}$, $\mathbf{W}_g \in \mathbb{R}^{d \times d}$, where $d$ is the feature size. By stacking such GCN layers, GCN can retrieve regional features for each node. 

In order to model the dependency relation type, we propose to use trainable latent features to represent each dependency relation type. Specifically, we preserve a trainable relational look-up table $\mathbf{R} \in \mathbb{R}^{|N| \times m}$, where $|N|$ is the number of dependency relation types and $m$ is the dependency relation feature size. Then, the novel \textsc{DreGcn} can be defined as:
\begin{equation}
\setlength{\abovedisplayskip}{5pt}
\setlength{\belowdisplayskip}{5pt}
  \label{dregcn}
  \begin{split}
   \mathbf{h}^{l+1}_{i} &= ReLU ( \sum_{j=1}^{n}\sum_{k=1}^{|N|}(\mathbf{A}_{ij}\mathbf{W}_r^{l+1}[\mathbf{h}^{l}_{j}; \mathbf{R}[k]]\mathbf{Q}_{ijk}+\mathbf{b}^{l+1}_r)
  \end{split}
\end{equation}
where [;] means concatenation, $\mathbf{W}_r \in \mathbb{R}^{d \times (d + m)}$, $\mathbf{Q}_{ijk}$ denotes whether there is the k-th dependency relation type between node $i$ and node $j$ or not. In doing so, the relational feature among nodes can be reasonably modeled and updated during training. 

\subsection{Task-specific Layers} 
For the opinion-passing, we make the information of opinion term available to the AS task, as done in~\cite{he_acl2019}. For the message-passing, we design a more effective mechanism for information sharing between multiple related tasks. Instead of passing the predictions of the AE task and the AS task shown in Eq.(\ref{eq:message-passing-predictions}), we propose to pass the original representation, which contains more abundant message than the probability distribution. The message-passing function is as follows:
\begin{equation}
\setlength{\abovedisplayskip}{5pt}
\setlength{\belowdisplayskip}{5pt}
  \label{eq:message-passing-representation}
  \begin{split}
   \mathbf{h}_i^{s(t)} &= f_{\theta _{re}}(\mathbf{h}_i^{s(t-1)};\mathbf{h}_i^{ae(t-1)};\mathbf{h}_i^{as(t-1)})
  \end{split}
\end{equation}
where $\mathbf{h}_i^{o(t-1)}$ ($o \in \{ae, as\}$) denotes the task-specific representation corresponding to $w_i$ after $t-1$ rounds of message-passing. The difference between the representation and the probability distribution is that the representation can be transformed to the probability by a fully-connected layer and a softmax layer. The new message-passing mechanism makes the rich information of the AE task and the AS task available to each other, and thus is more effective for the ABSA task, as empirically verified in the Ablation Study section. 

\subsection{Prediction} After $T$ times iteration, the predicted results for the AE task and the AS task are generated. Clearly, we can compute the score by directly counting the result for each task. To measure the overall performance, we need to obtain the aspect term-polarity pairs. Since the extracted aspect term may be composed of several tokens and the predicted polarities of each token may be inconsistent, we following~\cite{he_acl2019} only take the sentiment polarity of the first token of the current aspect term as the sentiment label.

\subsection{Training}

We simultaneously train the AE task and the AS task for message-passing. The loss function is as follows:
\begin{equation}
\setlength{\abovedisplayskip}{5pt}
\setlength{\belowdisplayskip}{5pt}
    \begin{split}
     \mathcal{J} = \frac{1}{N_a}\sum\limits_{i=1}^{N_a} \frac{1}{n_i} \sum\limits_{j=1}^{n_i} (min (- \sum_{k=0}^{C}\mathbf{y}_{i,j,k}^{ae} \log(\hat{\mathbf{y}}_{i,j,k}^{ae(T)})) + min (- \sum_{k=0}^{C}\mathbf{y}_{i,j,k}^{as} \log(\hat{\mathbf{y}}_{i,j,k}^{as(T)})))
    \end{split}
\label{aspect_loss}
\end{equation}
where $N_a$ denotes the total number of training instances, $n_i$ denotes the number of tokens contained in the $i$th training instance, $C$ is the class number, and $\mathbf{y}_{i,j,k}^{ae}$ ($\mathbf{y}_{i,j,k}^{as}$) denotes the ground-truth of the AE (AS) task. 
In all datasets, only aspect terms have sentiment annotations. We label each token which belongs to any aspect terms with the sentiment of the corresponding aspect terms. During training, we only consider AS predictions on these aspect term-related tokens for computing the AS loss and ignore the sentiments predicted on other tokens, i.e., $ce(\mathbf{y}_{i,j,k}^{as}, \hat{\mathbf{y}}_{i,j,k}^{as(T)}) = 0$ in Eq.(\ref{aspect_loss}) if $\mathbf{y}_{i,j,k}^{ae} \notin $\{\texttt{BA}, \texttt{IA}\}~\cite{he_acl2019}.


\section{Experiments}

\subsection{Datasets}
Table~\ref{datasets} shows the statistics of all datasets. We use three benchmark datasets, taken from SemEval 2014~\cite{Pontiki:14} and SemEval 2015~\cite{Pontiki:15}, to evaluate the effectiveness of our approach. The opinion terms are annotated by~\cite{rncrf}. We use {$\mathbb{D}_{\text{1}}$}, {$\mathbb{D}_{\text{2}}$}, and {$\mathbb{D}_{\text{3}}$} to denote SemEval-2014 Laptops, SemEval-2014 Restaurants, and SemEval-2015 Restaurants, respectively.

\renewcommand{\arraystretch}{1.1}
\begin{table}[!h]
\centering
\small
\scalebox{0.86}{
\setlength{\tabcolsep}{0.78mm}{
\begin{tabular}{llcccccc}
\toprule 
&\multirow{ 2}{*}{Datasets} & \multicolumn{3}{c}{Train} &  \multicolumn{3}{c}{Test}\\\cline{3-8}
&& Sentences & AT & OT & Sentences & AT & OT\\\hline
$\mathbb{D}_{\text{1}}$&Laptop14 &3,048 &2,373 &2,504 &800 &654 &674 \\
$\mathbb{D}_{\text{2}}$&Restaurant14 &3,044 &3,699 &3,484 &800 &1,134 &1,008\\
$\mathbb{D}_{\text{3}}$&Restaurant15 &1,315 &1,199 &1,210 &685 &542 &510\\
\bottomrule
\end{tabular}}}
\caption{Dataset statistics with numbers of sentences, aspect terms (AT) and opinion terms (OT).}\label{datasets}
\end{table}

\subsection{Experiment Settings}

\paragraph{Word Embeddings.} 
For general-purpose embeddings, we use \texttt{GloVe.840B.300d} 
released by~\cite{glove:14}. For domain-specific embeddings, we adopt the embeddings released by~\cite{xu_acl2018} as done in~\cite{he_acl2019}. 

\paragraph{Implementation Details.}
Our models\footnote{Code: https://github.com/XL2248/DREGCN} are trained by adam optimizer~\cite{Adam:14}, with the learning rate $\eta_0 = 0.0005$, and we set batch size to 50. 
At the training stage, as done in~\cite{he_acl2019}, we randomly sample 20\% of each training data as the development set and use the remaining 80\% only for training. More details are given in Appendix A. 
The tuning details about the layer number of GCN and the iteration $T$ are given in Appendix B.

\paragraph{Evaluation Metrics. }
We employ five metrics for evaluation and report the average score over 5 runs with random initialization in all experiments as done in~\cite{he_acl2019}. For the overall ABSA task, we compute the F1 score denoted as \textbf{F1-I} to measure the overall performance, where an extracted aspect term is taken as correct only when the span and the sentiment are both correctly identified. For the AE task, we use F1 to measure the performance of aspect term extraction and opinion term extraction, which are denoted as \textbf{F1-a} and \textbf{F1-o}, respectively. For the AS task, we adopt accuracy and macro-F1 to measure the performance of AS, which are denoted as \textbf{acc-s} and \textbf{F1-s}, respectively. The two metrics are computed based on the correctly extracted aspect terms from AE instead of the golden aspect term. 

\begin{table*}[t]
\centering
\small
\scalebox{0.94}{
\setlength{\tabcolsep}{0.64mm}{
\begin{tabular}{l|ccccc|ccccc|ccccc}
\toprule
\multirow{ 2}{*}{Methods} & \multicolumn{5}{c|}{$\mathbb{D}_{\text{1}}$} &  \multicolumn{5}{c|}{$\mathbb{D}_{\text{2}}$} & \multicolumn{5}{c}{$\mathbb{D}_{\text{3}}$}\\\cline{2-16}
& F1-a & F1-o & acc-s & F1-s & F1-I & F1-a & F1-o & acc-s & F1-s & F1-I & F1-a & F1-o & acc-s & F1-s & F1-I\\\hline
CMLA-ALSTM$^*$ &76.80 &77.33 &70.25 &66.67 &53.68    &82.45 &82.67 &{77.46} &68.70 &63.87     &68.55 &71.07 &81.03 &58.91 &54.79\\
CMLA-dTrans$^{*\dagger}$ &76.80 &77.33 &72.38 &69.52 &55.56    &82.45 &82.67 &{79.58} &72.23 &65.34     &68.55 &71.07 &82.27 &66.45 &56.09\\
DECNN-ALSTM$^*$ &78.38 &78.81 &70.46 &66.78 &55.05     &83.94 &85.60 &{77.79} &68.50 &65.26     &68.32 &71.22 &80.32 &57.25 &55.10\\
DECNN-dTrans$^{*\dagger}$ &78.38 &78.81 &73.10 &70.63 &56.60    &83.94 &85.60 &{80.04} &73.31 &67.25      &68.32 &71.22 &82.65 &69.58 &56.28\\
PIPELINE-IMN$^*$     &78.38 &78.81 &72.29 &68.12 &56.02    &83.94 &85.60 &{79.56} &69.59 &66.53      &68.32 &71.22 &82.27 &59.53 &55.96\\\hline
MNN$^*$  &76.94 &77.77 &70.40 &65.98 &53.80     &83.05 &84.55 &77.17 &68.45 &63.87      &70.24 &69.38 &80.79 &57.90 &56.57\\
INABSA$^*$ &77.34 &76.62 &72.30 &68.24 &55.88    &83.92 &84.97 &79.68 &68.38 &66.60      &69.40 &71.43 &82.56 &58.81 &57.38\\\hline
IMN$^{-d}$ wo DE$^*$ &76.96 &76.85 &72.89 &{67.26} &56.25     &83.95 &85.21 &79.65 &69.32 &66.96    &69.23 &{68.39} &81.64 &57.51 &56.80 \\
IMN$^{-d*}$&78.46 &78.14 &73.21 &{69.92} &57.66     &84.01 &85.64 &81.56 &71.90 &68.32     &69.80 &{72.11} &83.38 &60.65 &57.91 \\
IMN$^{*\dagger}$ &77.96 &77.51 &75.36 &72.02 &{58.37}    &{83.33} &{85.61} &{83.89} &{75.66} &{69.54}    &{70.04} &71.94 &{85.64} &{71.76} &{59.18}\\
IMN$^{*\dagger}$+BERT &78.47 &79.05 &77.18 &74.56 &{60.53}    &{85.22} &{86.64} &\bf{84.90} &\bf{76.54} &{71.33}    &{72.55} &72.43 &{84.37} &{71.28} &{60.76}\\\hline
\textsc{DreGcn} wo DE (Ours) &76.30 &73.92 &75.83 &{71.05} &57.48     &83.75 &84.09 &80.78 &71.23 &67.51    &{68.63} &70.09 &84.25 &71.29 &57.70 \\
\textsc{DreGcn} (Ours) &77.78 &76.62 &77.18 &72.27 &{59.66}     &{84.16} &{85.04} &81.27 &{72.48} &{68.94}     &{69.36} &70.75 &{86.03} &{66.89} &{59.71}\\
\textsc{DreGcn}+CNN (Ours) &{79.45} &{75.40} &{77.86} &{73.46} &{61.60}     &{85.93} &{86.05} &{81.88} &{73.32} &{70.21}    &{71.00} &{70.55} &\bf{86.16} &\bf{73.35} &{61.06} \\\cline{2-16}
\textsc{DreGcn}+CNN+BERT(Ours) &\bf{79.78} &\bf{79.21} &\bf{79.37} &\bf{76.37} &\bf{63.04}   &\bf{87.00} &\bf{86.95} &{83.61} &{75.79} &\bf{72.60}    &\bf{73.30} &\bf{72.60} &{85.25} &{73.02} &\bf{62.37}\\
\bottomrule
\end{tabular}}}
\caption{Model comparison. The results with ``$^*$'' are retrieved from IMN~\cite{he_acl2019}. ``$\dagger$'' represents that these models utilize a large document-level corpus. ``$^{-d}$'' denotes without using document-level corpus. ``wo DE'' indicates without using domain-specific embeddings. ``$+BERT$'' denotes exploiting {\em BERT-BASE} features on ``\textsc{DreGcn}+CNN''. In the Encoder Layers of Figure~\ref{fig:architecture}, ``IMN$^{-d}$'' means only {\em CNN$s^s$ module}, and ``\textsc{DreGcn}'' means only \emph{DREGCN} module. The results of ours do not use any document-level corpus.  
}\label{main results}
\end{table*}
\subsection{Compared Models}
\begin{itemize}
    \item \noindent\textbf{Pipeline Approach.}
    
    \{\textbf{CMLA}, \textbf{DECNN}\}-\{\textbf{ALSTM}, \textbf{dTrans}\}: The four methods are constructed by two best-performing models for two subtasks. For AE task, we select CMLA~\cite{wang2017coupled} and DECNN~\cite{xu_acl2018}. The former is proposed for the AE task through modeling their inter-dependencies. The latter utilizes a multi-layer CNN structure as encoder with double embeddings. For AS task, ATAE-LSTM (denoted as ALSTM for short)~\cite{Wang:16} and the model from~\cite{He:18} (denoted as dTrans) are used. ALSTM is an attention-based LSTM structure. The dTrans introduces a large document-level corpus to improve the AS performance.
    
    \textbf{PIPELINE-IMN}: It means the pipeline setting of IMN~\cite{he_acl2019}, which trains the AE task and the AS task separately. 
    
    \textbf{SPAN-pipeline}~\cite{hu-etal-2019-open}: This work investigates those three methods (i.e. pipeline, integrated and joint) with BERT as backbone networks, which obtains the best results with \textbf{SPAN-pipeline} method. We replace BERT-Large with BERT-Base in their released code to get the result.
    \item \noindent\textbf{Integrated Approach.} 
    
    \textbf{MNN}~\cite{Wang18}: It handles this task as a sequence labeling task with a unified tagging scheme.
    
    \textbf{INABSA}~\cite{li2019unified}: This model leverages a unified tagging scheme to integrate the two subtasks of ABSA.
    
    \textbf{BERT+GRU}~\cite{li-etal-2019-exploiting}: It explores the potential of BERT for ABSA task.
    
    \item \noindent\textbf{Joint Approach. }
    
    \textbf{DOER}~\cite{Luo2019doer}:  This model employs a cross-shared unit to jointly train the two subtasks.
    
    \textbf{IMN}~\cite{he_acl2019}: It is the current state-of-the-art method, which uses an interactive architecture with multi-task learning for end-to-end ABSA task. {``IMN}$^{-d}$ wo DE'' and ``{IMN}$^{-d}$'' are the variants of {IMN}.

\end{itemize}




\begin{table*}[t]
\centering
\begin{minipage}{0.45\linewidth}
\small
\scalebox{0.9}{
\setlength{\tabcolsep}{0.9mm}{
\begin{tabular}{clc}
\toprule 
Row&Model  &$\mathbb{D}_{\text{1}}$\\\hline
0&DOER~\cite{Luo2019doer}$^{\natural}$ &59.48   \\
1&\textsc{DreGcn}+CNN (Ours)  &\bf{61.60}   \\\hline
2&BERT+GRU ($BERT_{BASE}$)~\cite{li-etal-2019-exploiting}$^{\natural}$ &60.42   \\
3&SPAN-pipeline ($BERT_{BASE}$)~\cite{hu-etal-2019-open}$^{\natural}$ &61.84   \\
4&\textsc{DreGcn}+CNN+$BERT_{BASE}$ (Ours) &\bf{63.04}  \\
\bottomrule
\end{tabular}}}
\caption{F1-I (\%) scores on $\mathbb{D}_{\text{1}}$, which is our common dataset. ``$^{\natural}$'' indicates that the results are generated by running their released code under our experimental setting (dataset). 
}\label{smallexperiment}
\end{minipage}
\qquad\quad
\begin{minipage}{0.45\linewidth}
\centering
\small
\scalebox{0.86}{
\setlength{\tabcolsep}{0.5mm}{
\begin{tabular}{clccc}
\toprule 
Row&Model  &$\mathbb{D}_{\text{1}}$&$\mathbb{D}_{\text{2}}$&$\mathbb{D}_{\text{3}}$\\\hline
0&\textsc{CNN} &56.66  &66.32  &{57.91} \\
1&Vanilla GCN (Eq.(\ref{gcn})) &57.10 &65.00 &{56.86}  \\
2&\textsc{DreGcn} (Eq.(\ref{dregcn})) &57.46  &66.25  &{58.32} \\
3&+Opinion-passing (Eq.(\ref{revelance_fun}))  &57.89  &66.51 &{58.57}  \\
4&+Message-passing predictions (Eq.(\ref{eq:message-passing-predictions})) &58.50 &67.36 &{57.92}  \\
5&+Message-passing representations (Eq.(\ref{eq:message-passing-representation})) &61.60  &70.21  &{61.06}  \\
\bottomrule
\end{tabular}}}
\caption{F1-I (\%) scores of ablation study. The component (i.e., Rows 3$\sim$5) is added on the \textsc{DreGcn} (i.e., Row 2), respectively.}\label{ablation}
\end{minipage}
\end{table*}
\subsection{Results and Analysis}
\paragraph{Overall Performance.}
Table~\ref{main results} and Table~\ref{smallexperiment} present the results of our models and baseline models for the complete ABSA task. 
Results show that our model consistently outperforms all baseline models by a large margin on all datasets in most cases even without BERT. Since there is no syntax-based method for the overall ABSA task to compare with, we also conduct experiments on the separate subtask setting, i.e, the AE and AS task, which are presented in Appendix D. 
From Table~\ref{main results} and Table~\ref{smallexperiment}, we can conclude:

1) For the overall performance (F1-I), Table~\ref{main results} shows that ``\textsc{DreGcn}+CNN'' is able to significantly surpass other baselines. Concretely, ``\textsc{DreGcn}+CNN'' outperforms the best F1-I results of IMN by \textbf{3.23\%}, \textbf{0.67\%}, and \textbf{1.88\%} on $\mathbb{D}_{\mathrm{1}}$, $\mathbb{D}_{\mathrm{2}}$, and $\mathbb{D}_{\mathrm{3}}$, respectively\footnote{Note that our approach does not use any document-level corpus, while IMN exploits this additional corpus.}, suggesting that \textsc{DreGcn} and message-passing mechanism have an overall positive impact on the ABSA task. We notice that the improvement of our method on $\mathbb{D}_{\mathrm{2}}$ is marginal by contrast with IMN. The reason may be that $\mathbb{D}_{\mathrm{2}}$ contains a large number of ungrammatical sentences (14.3\%), which affect the accuracy of dependency parsing. After using $BERT_{BASE}$ features, we achieve further improvements (\textbf{+4.67\%}, \textbf{+3.06\%}, and \textbf{+3.19\%} compared with IMN, respectively). Besides, the results also show that domain-specific knowledge is very helpful (``IMN$^{-d}$ wo DE'' vs. IMN$^{-d}$ and ``\textsc{DreGcn} wo DE'' vs. \textsc{DreGcn}). 

2) For AE (F1-a and F1-o in Table~\ref{main results}), ``\textsc{DreGcn}+CNN'' performs the best in most cases than baselines. Those results demonstrate the effectiveness of our model, which indeed benefits from the dependency structure information and message-passing mechanism. This shows that the syntax information is very pivotal to the AE task. 

3) For AS (acc-s and F1-s in Table~\ref{main results}), even though some methods (IMN and the pipeline methods with dTrans) utilize additional knowledge by joint training with document-level tasks, \textsc{DreGcn} still significantly surpasses the baseline methods. This suggests that our model can sufficiently model the dependency structure and indeed benefit from the message-passing mechanism. This shows that the syntax information is crucial for the AS task. 

4) Table~\ref{smallexperiment} shows the results of our model and another strong baselines: DOER, ``BERT+GRU'' and SPAN-pipeline. We find that ``\textsc{DreGcn}+CNN'' can surpass DOER and even be highly comparable with BERT-based models. Our model with BERT (Row 4) can also outperform the ``BERT+GRU'' (Row 2) and SPAN-pipeline (Row 3), which suggests the effectiveness of our proposed approach. Besides, we investigate the impact of BERT CLS at different positions in the model, which are given in Appendix C.

\paragraph{Ablation Study.} 
To investigate the impact of different components, we conduct ablation studies in Table~\ref{ablation}, where Rows 1$\sim$2 are conducted without any informative message-passing, and add other components on \textsc{DreGcn} one at a time (Rows 3$\sim$5). From Table~\ref{ablation}, we can conclude:
\begin{enumerate}[1).]\setlength{\itemsep}{0.05cm}
\item 
Considering dependency relation types as features between nodes is helpful with considerable performance gains to the ABSA task (Row 2 vs. Row 1 \& Row 0), which shows that the syntax information is very critical for both aspect term extraction and sentiment recognition.  
\item 
Opinion message can indeed help the AS task and thus improves the overall performance (Row 3 vs. Row 2).
\item  Message-passing makes a large contribution to the overall performance (Row 4 \& Row 5 vs. Row 2).
\item  
Transferring representations (our proposed message-passing mechanism) is more helpful than passing predictions (Row 5 vs. Row 4), which is intuitive that original representations have richer information than the probability distribution. 
\end{enumerate}

\paragraph{Case Study.}
To provide an intuitive understanding of how the \textsc{DreGcn} works, we present some examples in Table~\ref{case_study}. As observed in Example 1 and 2, the ``Vanilla GCN'' correctly predicts the opinion term and the sentiment while it fails to produce the right aspect term. With the help of modeling the dependency relation type: $windows\xrightarrow{nummod}8$ and $size\xrightarrow{conj}speed$ (i.e. by a coordinating
conjunction word $and$), \textsc{DreGcn} can correctly handle these two cases, which suggests that dependency relation type is indeed critical to the AE task. For the sentiment orientation of multiple aspect terms, our model is not confused when identifying the sentiment polarity in Example 3. Here, \textsc{DreGcn} can accurately predict the sentiment polarity because of modeling the dependency relation. 
For Example 4, since no opinion word is mentioned in this sentence, ``\emph{\textbf{device}}'' should not be regarded as an aspect term. \textsc{DreGcn} avoids to extract this kind of terms by aggregating information from rich opinion and sentiment representation, which demonstrates the effectiveness of our message-passing mechanism. For Example 5, due to combining the dependency relation type: $veal\xrightarrow{conj}mushrooms$ with the message-passing mechanism, \textsc{DreGcn} correctly handles this case even though ``\emph{\textbf{veal}}'' is an uncommon word in the training corpus.

\begin{table*}[ht]
    \centering
    \scalebox{0.7}{
    {%
    \begin{tabular}{@{}L{7.6cm}@{~}|@{~}L{1.6cm}@{~}|@{~}L{2.9cm}@{~}|@{~}L{1.7cm}@{~}|@{~}L{3.2cm}@{~}|@{~}L{1.7cm}@{~}|@{~}L{2.7cm}@{~}}
    \Xhline{3\arrayrulewidth}
        \multirow{2}{*}{Examples (Golden labels are marked.)} & \multicolumn{2}{c|@{~}}{Vanilla GCN} & \multicolumn{2}{c|@{~}}{IMN} & \multicolumn{2}{c@{~}}{\textsc{DreGcn}}  \\ \cline{2-7}
        &Opinion &Complete &Opinion &Complete &Opinion &Complete   \\ \hline
        1. Biggest \textcolor{blue}{\emph{complaint}} is [\textcolor{red}{windows 8}]$_{\text{pos}}$ & complaint &[windows]$_{\text{pos}}$(${\text{\xmark}}$) &complaint &[windows 8]$_{\text{pos}}$&complaint &[windows 8]$_{\text{pos}}$ \\\hline
        2. It is the \textcolor{blue}{\emph{perfect}} \textcolor{red}{[size]}$_{\text{pos}}$ and \textcolor{red}{[speed]}$_{\text{pos}}$  for me. & perfect &[size]$_{\text{pos}}$, None (${\text{\xmark}}$) &perfect  &[size]$_{\text{pos}}$,[speed]$_{\text{neu}}$(${\text{\xmark}}$) &perfect &[size]$_{\text{pos}}$,[speed]$_{\text{pos}}$\\\hline
        3. \textcolor{red}{[Coffee]}$_{\text{pos}}$ is a \textcolor{blue}{\emph{better}} deal than \textcolor{blue}{\emph{overpriced}} \textcolor{red}{[cosi sandwiches]}$_{\text{neg}}$  & better, None (${\text{\xmark}}$) &[Coffee]$_{\text{pos}}$, [sandwiches]$_{\text{pos}}(${\text{\xmark}}$)$ &better, overpriced &[Coffee]$_{\text{pos}}$, [cosi sandwiches]$_{\text{pos}}(${\text{\xmark}}$)$ &better, overpriced &[Coffee]$_{\text{pos}}$, [cosi sandwiches]$_{\text{neg}}$ \\\hline
        4.The device speaks about itself. & None &[device]$_{\text{pos}}$(${\text{\xmark}}$) &None &[device]$_{\text{neu}}$(${\text{\xmark}}$) &None &None \\\hline 
        5. The \textcolor{red}{[veal]}$_{\text{pos}}$ and the \textcolor{red}{[mushrooms]}$_{\text{pos}}$ were cooked \textcolor{blue}{\emph{perfectly}}. & perfectly &None(${\text{\xmark}}$), [mushrooms]$_{\text{pos}} $ &perfectly &[veal]$_{\text{neu}}$(${\text{\xmark}}$), [mushrooms]$_{\text{pos}}$ &perfectly &[veal]$_{\text{pos}}$, [mushrooms]$_{\text{pos}}$\\
        \Xhline{3\arrayrulewidth}
    \end{tabular}}}
    \caption{Case study. The ``Opinion'' and ``Complete'' columns denote the opinion terms and aspect terms with corresponding sentiment polarities, respectively.  ``${\text{\xmark}}$'' indicates incorrect predictions. }
    \label{case_study}
\end{table*}
\section{Related Work}
\paragraph{Aspect-based Sentiment Analysis.}
There are two sub-tasks in ABSA, namely, the aspect term extraction task~\cite{qiu2011opinion,dtbcsnn,he-etal-2017-unsupervised,rncrf,wang2017coupled,wdbem,yin2019pod,li2017deep,li2018aspect,angelidis-lapata-2018-summarizing,fan-etal-2019-target,ma-etal-2019-exploring}
and the aspect-level sentiment classification task~\cite{vo2015target,Tang:16a,tang-etal-2019-progressive,Zhang:2016:GNN,Wang:16,Wang:2019:ASA:3308558.3313750,wang-etal-2019-learning-noisy,wang-etal-2019-investigating,wang-etal-2019-aspect,Liu:17,Chen:17,chen-qian-2019-transfer,ma2018targeted,li2018transformation,li-etal-2019-transferable,hu-etal-2019-constrained,li-lu-2019-learning,du-etal-2019-capsule,bao-etal-2019-attention,yang2019aspect,sun-etal-2019-utilizing,liang-etal-2019-novel-aspect,jiang-etal-2019-challenge}, 
which have been deeply studied as two separate tasks in the past. Recently, some methods attempt to solve the overall ABSA task simultaneously. Concretely, a unified tagging scheme is applied to address it as a sequence labeling task, while the inter-dependency relation between the two tasks is not explicitly modeled~\cite{zhang-etal-2015-neural,Wang18,li2019unified}. Therefore, some studies propose to take them as two sequence tagging tasks and jointly model them, which generate some promising results in this direction~\cite{he_acl2019,Luo2019doer}. 
However, the syntax information is not considered in their models, which is important to the ABSA task. Although some work involves the syntax information in separate subtask settings, they do not sufficiently exploit that information to enhance the overall ABSA task. For example,  Dong et al.~\shortcite{dong-etal-2014-adaptive} and  Nguyen and Shirai~\shortcite{nguyen-shirai-2015-phrasernn} need to convert the dependency structure into a binary tree and then adjust the target aspect as the root node, which may lead to the opinion word far away from the target aspect, while GCN can overcome the limitation over the original order of the dependency graph~\cite{huang-carley-2019-syntax,peng2019knowing,zhang-etal-2019-aspect}.
\paragraph{Graph Convolutional Network.}
GCN~\cite{kipf2017semi} has been extensively studied in many natural language processing (NLP) tasks. 
The ABSA task is no exception, for instance, most existing GCN-based methods are for separate AS task setting, in which  Zhao et al.~\shortcite{DBLP:journals/corr/abs-1906-04501} focus on modeling the sentiment dependencies over multiple aspect terms in one sentence; Sun et al.~\shortcite{sun-etal-2019-aspect}, Huang and Carley~\shortcite{huang-carley-2019-syntax}, Hou et al.~\shortcite{hou2019selective} and Zhang et al.~\shortcite{zhang-etal-2019-aspect} focus on encoding more aspect-specific representations by using vanilla GCN (GAN) on dependency graphs without considering dependency types. 

Different from all studies above, in this work, we focus on the end-to-end ABSA task and extend vanilla GCN through embedding dependency types into the model for capturing more fine-grained linguistic knowledge (i.e. the dependency relation and type) at the relational level in a joint framework, and obtain better performances.

\section{Conclusions}

In this paper, we propose a dependency syntactic knowledge augmented interactive architecture for end-to-end ABSA task, which can fully exploit the syntax information through a well-designed dependency relation embedded graph convolutional network (\textsc{DreGcn}) and jointly model multiple related tasks. In addition, we design a more effective message-passing mechanism to enable our model to learn information representation from multiple tasks. The experimental results on three benchmark datasets demonstrate the effectiveness of our proposed approach, which achieves new state-of-the-art results. Besides, using BERT as an additional feature extractor, we provide further improvements. 

\section*{Acknowledgements}
Liang, Xu and Chen are supported by the National
Natural Science Foundation of China (Contract
61370130, 61976015, 61473294 and 61876198),
and the Beijing Municipal Natural Science Foundation (Contract 4172047), and the International
Science and Technology Cooperation Program
of the Ministry of Science and Technology
(K11F100010).


\end{document}